\pdfoutput=1

\documentclass[11pt]{article}

\usepackage{ACL2023}

\usepackage{times}
\usepackage{latexsym}
\usepackage{multicol}
\usepackage{multirow}
\usepackage{CJKutf8}
\newcommand{\chinese}[1]{%
  \begin{CJK}{UTF8}{gbsn}#1\end{CJK}%
}
\usepackage{amsmath}
\usepackage{booktabs}
\usepackage{graphicx}
\usepackage{tcolorbox}
\usepackage{algorithm}
\usepackage{algpseudocode}  
\tcbuselibrary{breakable}
\usepackage{xcolor}
\usepackage[T1]{fontenc}
\algrenewcommand{\algorithmicrequire}{\textbf{Input:}} 
\algrenewcommand{\algorithmicensure}{\textbf{Output:}}  

\usepackage[utf8]{inputenc}

\usepackage{microtype}

\usepackage{inconsolata}

\title{RAIR: Retrieval-Augmented Iterative Refinement for Chinese Spelling Correction}

\author{Junhong Liang$^{1,2}$, Yu Zhou$^{1,3}$\\
$^{1}$Institute of Automation, Chinese Academy of Sciences \\
 $^{2}$School of Artificial Intelligence, University of Chinese Academy of Sciences \\ $^{3}$Fanyu AI Laboratory, Zhongke Fanyu Technology Co., Ltd, Beijing, China\\
  \texttt{liangjunhong2022@ia.ac.cn}}

\begin{document}
\maketitle
\begin{abstract}
Chinese Spelling Correction (CSC) aims to detect and correct erroneous tokens in sentences. Traditional CSC focuses on equal length correction and uses pretrained language models (PLMs). While Large Language Models (LLMs) have shown remarkable success in identifying and rectifying potential errors, they often struggle with adapting to domain-specific corrections, especially when encountering terminologies in specialized domains. To address domain adaptation, we propose a \textbf{R}etrieval-\textbf{A}ugmented \textbf{I}terative \textbf{R}efinement (RAIR) framework. Our approach constructs a retrieval corpus adaptively from domain-specific training data and dictionaries, employing a fine-tuned retriever to ensure that the retriever catches the error correction pattern. We also extend equal-length into variable-length correction scenarios. Extensive experiments demonstrate that our framework outperforms current approaches in domain spelling correction and significantly improves the performance of LLMs in variable-length scenarios. 
\end{abstract}

\section{Introduction}

Chinese Spelling Correction (CSC) is a long-established task aimed at correcting misspelled characters in a sentence.  Traditional CSC tasks focus on correcting general texts, such as errors in daily writing \citep{xu_read_2021,hu2022cscd},  Optical Character Recognition (OCR) \citep{zhang2015characterlevel}, or texts produced by Chinese learners \citep{tseng-etal-2015-introduction}. However, Chinese spelling errors also appear in domain-specific scenarios \citep{song_2023_rspell, wu2023rethinking}. Addressing domain-specific spelling errors is crucial because the precision and accuracy of domain documents are essential for communication and avoiding misunderstanding. Traditional CSC methods are based on pre-trained language models (PLMs) such as BERT structure \cite{zhang2020spelling,xu_read_2021} to capture the inherent semantic information of each token, while the complex structure and meticulous design of knowledge fusion of PLMs inherently narrow the scope in addressing domain-specific spelling errors, resulting in less effective results. 

\begin{figure}[!t]
    \centering
    \includegraphics[width=\linewidth]{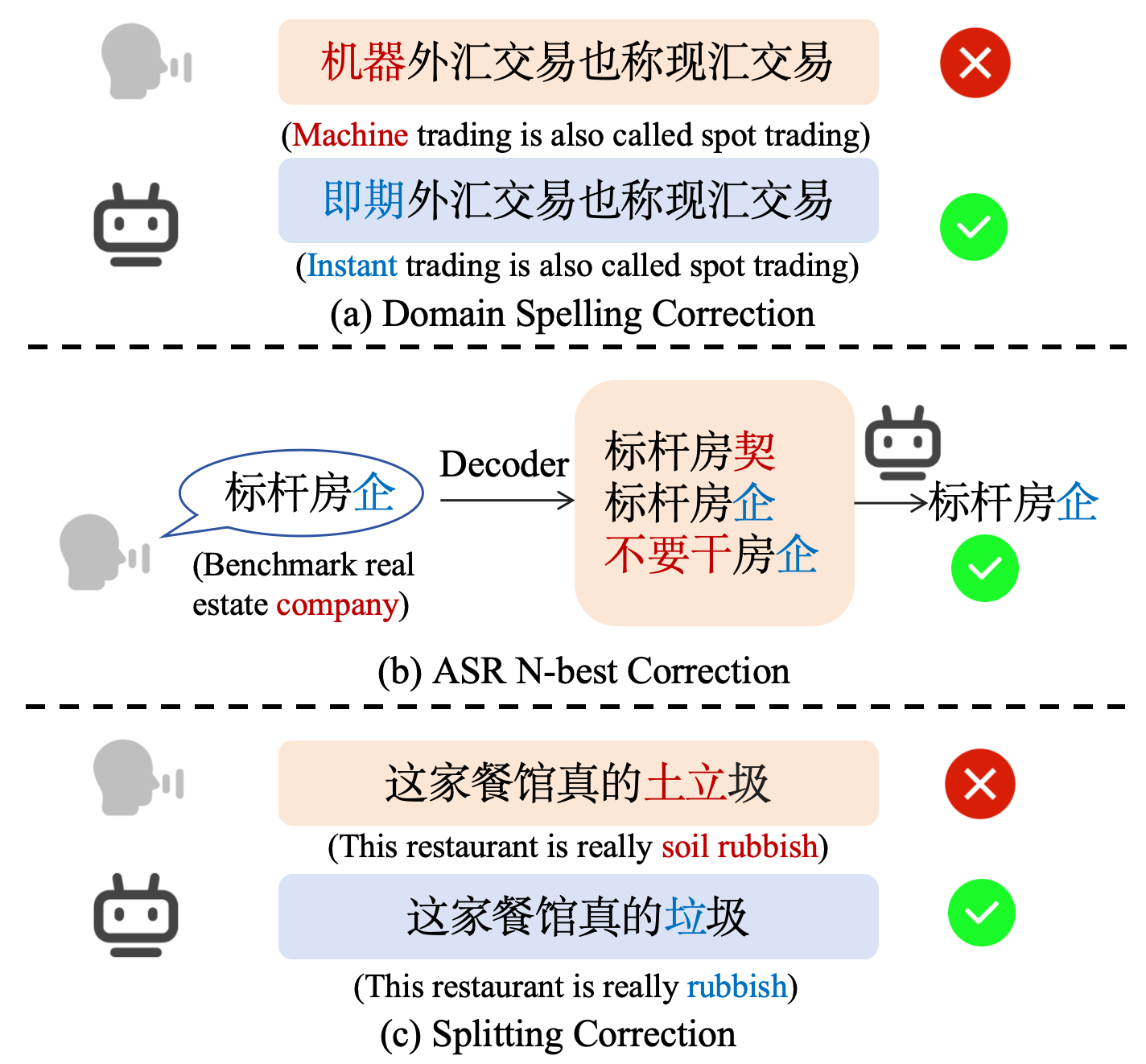}
    \caption{Examples of Chinese error correction in different scenarios: a) Domain Spelling Correction, where an LLM must rectify semantically incorrect spelling from domain data while preserving sentence length. b) ASR N-best Correction, in which the LLM combines multiple speech recognition hypotheses and outputs the correct sentence from the speaker, and c) Splitting Correction, requiring the LLM to restore erroneously split characters into a single character. The wrong characters and their corrections are marked in \textcolor{red}{red}/\textcolor{blue}{blue} respectively (best viewed in color).}
    \label{fig:csc_example}
\end{figure}

LLMs demonstrated excellent performance in several NLP tasks, and using LLMs for CSC has become a research trend. Most research introduces LLM as a generator to correct the equal-length spelling errors, such as by redefining tokenization rules \cite{li2024cllmlearncheckchinese} or providing character information using in-context learning \cite{dong2024rich}. However, challenges still exist for LLM methods. Firstly, it is essential to employ domain-specific knowledge to deal with spelling errors in specific domains, whereas existing approaches often overlook domain-specific insights. Secondly, it is vital to maintain the equal-length requirement in domain spelling correction, while the auto-regressive generation pattern of LLMs may fail to comply with strict length constraints in CSC tasks. Lastly, variable-length correction scenarios, such as ASR correction and character splitting, also exist in daily life and are insufficiently investigated by current spelling correction methods. The investigated error types are shown in Figure \ref{fig:csc_example}. 

These observations lead to a fundamental question overlooked by previous studies: \textit{\textbf{Can we utilize the generation abilities in LLMs to solve equal-length domain spelling errors while addressing variable length scenarios?}}

To address the issues mentioned above, we propose the RAIR (\textbf{R}etrieval \textbf{A}ugmented \textbf{I}terative \textbf{R}efinement) framework based on retrieval-augmented generation (RAG). This framework not only handles domain spelling errors but also ASR N-best Errors and Splitting Errors. We first dynamically construct a retrieval corpus from various sources, then fine-tune the semantic search retriever to catch the error correction pattern, and incorporate Multi-turn Length Reflection (MLR) to guide the output format for equal and variable length scenarios, finally using adaptive selection to combine correction prediction from direct and retrieval augmented correction. Our contributions are summarized as follows:

\begin{itemize}
    \item We present a novel plug-and-play CSC framework using RAG for domain spelling errors, which integrates retrieved correct sentences into the context of LLMs, along with length reflection employing iterative reasoning steps to ensure the correct output length.
    
    \item We extend the current equal-length domain spelling correction into variable-length correction scenarios. We propose a domain-adaptive retrieval dataset construction method that synthesizes multi-source data, allowing adaptive data construction for different domains and error types. 
    
    \item Extensive experiments demonstrate our framework outperforms current approaches in domain spelling correction and significantly improve the performance of LLMs in variable-length scenarios. 
\end{itemize}

\section{Related Work}

\subsection{Chinese Error Correction Datasets}
Traditional CSC datasets such as SIGHAN13/14/15 focus on the errors made by Chinese learners \cite{wu-etal-2013-chinese,yu-etal-2014-overview,chang-etal-2015-introduction}; however, these data are unable to reflect the errors that exist in native speakers. Therefore, \cite{hu2022cscd} proposes CSCD-NS, which collects the spelling errors in social media. In order to explore domain error correction abilities, \citet{lv_general_2023} creates ECSpell datasets, creating domain-specific errors in three domains, and \citet{wu2023rethinking} proposes the LEMON dataset, including real spelling errors in seven domains. To investigate further variable-length correction, several datasets such as CSEC \cite{liang-etal-2024-hybrid} and ChineseHP \cite{tang_pinyin_2024} are proposed to address the splitting and ASR decoding errors.

\subsection{Chinese Spelling Correction Methods}

Traditional CSC approaches primarily employ transformer-based PLMs, which can be categorized into two main frameworks. The Information-Learning architecture integrates phonetic and visual glyph features for direct error correction \citep{hong_faspell_2019,sun-etal-2021-chinesebert,xu_read_2021}. The Detection-Correction architecture adopts a two-stage process, first identifying errors, then correcting them \citep{zhang2020spelling,zhu-etal-2022-mdcspell,wu2023rethinking}. Recent advancements have merged these approaches, such as combining Chinese splitting features with soft-masked prediction \citep{liang-etal-2024-hybrid}, and improving BERT through phonetic-aware pretraining \citep{feng2025cnmbertmodelhanyupinyin}.

The advent of LLMs has significantly advanced Chinese spelling correction research. Recent work has systematically evaluated LLMs' error correction capabilities \cite{li2023ineffectivenesslargelanguagemodels,liang2024perlpinyinenhancedrephrasing}, and there exist several innovative approaches. Prompt engineering techniques have efficiently integrated semantic information into the context of LLMs  \citep{dong2024rich}. Besides, pronunciation and glyph similarity techniques have proven valuable for output probability calibration \citep{zhou2024simpleeffectivetrainingfreepromptfree}.
while multimodal methods incorporating speech-text alignment enhance error localization \cite{yuhang2024bridgingspeechtextenhancing}. Model architecture innovations have also shown promise, particularly through character-level tokenization \citep{li2024cllmlearncheckchinese} and specialized error detection training \citep{xu2024enhancingcharacterlevelunderstandingllms}.

\begin{figure*}[!t]
    \centering
    \includegraphics[width=1\linewidth]{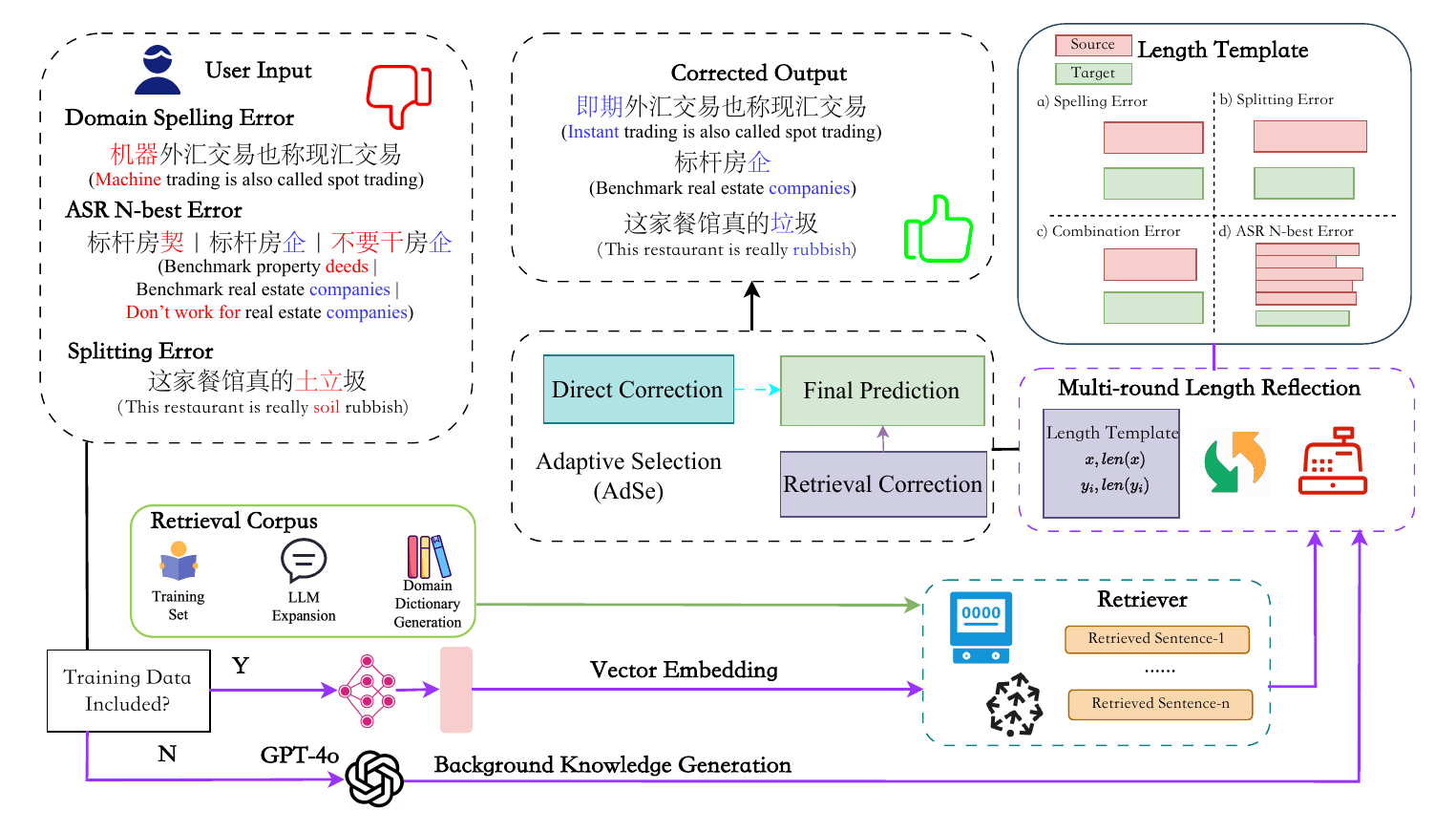}
    \caption{Structure diagram of RAIR, including several modules such as retrieval corpus, retriever, multi-round length reflection and adaptive selection (best viewed in color).}
    \label{fig:rair_framework}
\end{figure*}

Apart from this, notable advancement comes from RAG approaches. Current solutions demonstrate effectiveness through two key mechanisms: error-robust retrieval that utilizes training data as knowledge sources \citep{yin2024errorrobustretrievalchinesespelling}, and few-shot corpus retrieval\citep{dong-etal-2025-retrieval}.

\section{RAIR Framework}

Our proposed framework is displayed in Figure \ref{fig:rair_framework}, including a retrieval corpus, fine-tuned retriever, multi-turn length reflection, and adaptive selection.

\subsection{Problem Definition}

\paragraph{Existing Definition of Spelling Correction}
Given an input sentence $\boldsymbol{X} = [x_1,\ldots,x_n]$ consisting of $n$ characters, the goal is to generate a corresponding correct sentence $\boldsymbol{Y}$. If $\boldsymbol{X}$ contains a misspelled character $x_i$ with its correct form being $x_i'$, then the output should be $\boldsymbol{Y} = [x_1,\ldots,x_{i-1},x_i',x_{i+1},\ldots,x_n]$. This task requires the input and output lengths to remain consistent.

Traditional approaches focus only on equal-length spelling error correction; we extend the concepts into variable-length correction.

\paragraph{Definition of Splitting Correction} 
Given an input sentence $\boldsymbol{X} = [x_1,\ldots,x_m]$ consisting of $m$ characters, where there may exist a continuous character sequence $x_i,\ldots,x_{i+l-1}$ of length $l$ formed by splitting a single Chinese character. The original character $\hat{x}$ can be restored by merging these characters. The objective is to generate an output sentence $\boldsymbol{Y} = [x_1,\ldots,x_{i-1},\hat{x},x_{i+l},\ldots,x_n]$. Since this task involves character merging, the input and output lengths may differ.

\paragraph{Definition of ASR N-best Correction}
Given a set of candidate sentences from the ASR decoder $\boldsymbol{X} = (X_1, X_2, \ldots, X_n)$, where $X_i$ represents the $i$-th candidate sentence with length $l_i$, the goal is to generate the correct target sentence $\boldsymbol{Y}$.


\subsection{Creation of Retrieval Corpus}  
For each dataset, we construct the retrieval corpus from three perspectives.

\begin{itemize}
    \item \textbf{Domain Terms with Explanations}: Utilizing the THUOCL\footnote{\url{http://thuocl.thunlp.org/}} lexicon, domain-specific terms from the legal and medical fields are extracted and denoted as $W_D$. These terms are then processed by GPT-4o as a domain expert $E$ to generate detailed explanations for each term. This part is represented as $E(W_D)$.  
    \item \textbf{Training Set}: Using the training set samples, the correct sentence from each correct-incorrect sentence pair is obtained and denoted as $S_{\text{train}}$.
    \item \textbf{Training Set Expansion}: For each sentence in $S_{\text{train}}$, GPT-4o is used to generate a coherent paragraph based on the sentence. These paragraphs are then split into individual sentences, forming another part of the retrieval corpus, denoted as $E(S_{\text{train}})$. 
\end{itemize}

The final retrieval corpus $S_R$ is the union of these three parts:  
\[
S_R = \{S_{\text{train}},E(S_{\text{train}}), E(W_D)\}.
\]
For datasets lacking training data, the generative capability of LLMs like GPT-4o is leveraged to construct background descriptions for each test source sentence. Each generated paragraph then serves as the retrieval result for its corresponding sentence, ensuring no data leakage occurs.

\begin{figure}[!h]
  \includegraphics[width=1\columnwidth]{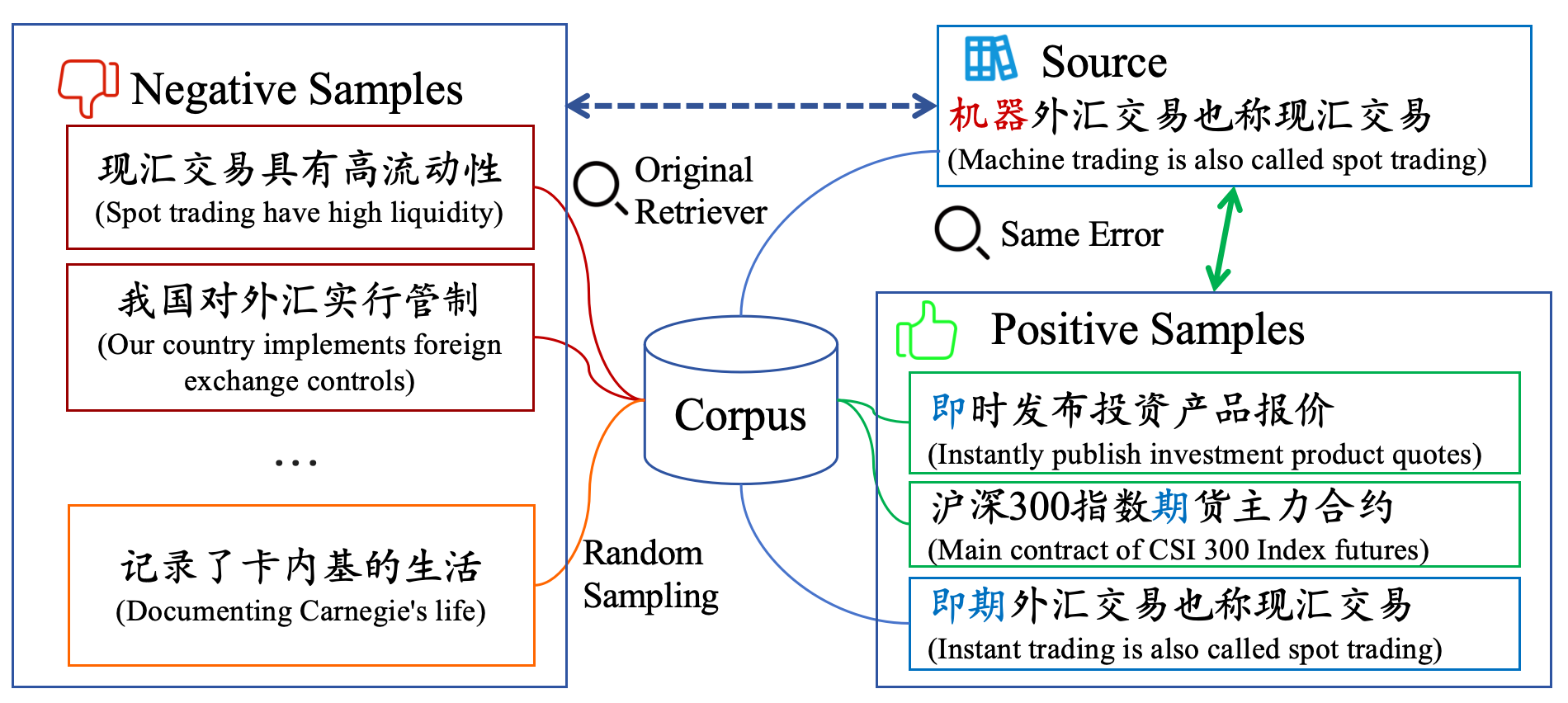}
  \caption{Construct positive and negative samples of retrievers}
  \label{fig:retrieval_method}
\end{figure}

\subsection{Retrieval Module}

The base retrieval model \( M_{\text{Ret}} \) extracts the top-\( k \) most semantically relevant sentences \( \{r_1, r_2, \dots, r_k\} = M_{\text{Ret}}(x) \) for a given source sentence \( x \). However, when \( x \) contains erroneous characters (e.g., ``\chinese{\textcolor{red}{机器}外汇交易也称现汇交易}''), \( M_{\text{Ret}} \) may fail to retrieve relevant results due to its dependency on character-level labels for sentence vectorization. 

To enhance robustness, as illustrated in Figure~\ref{fig:retrieval_method}, we fine-tune \( M_{\text{Ret}} \) via supervised learning using carefully constructed training samples. These samples consist of positive and negative examples. Positive samples include: (1) the correct target sentence \( y \) (e.g., ``\chinese{\textcolor{blue}{即期}外汇交易也称现汇交易}''), and (2) corpus sentence with the same corrected character (e.g., ``\chinese{\textcolor{blue}{即}时发布产品投资报价}'' and ``\chinese{沪深300指数\textcolor{blue}{期}货主力合约}''), where the source sentence incorrectly replaces \chinese{\textcolor{blue}{即期}} (instant) with \chinese{\textcolor{red}{机器}} (machine). Negative samples comprise: (1) semantically related sentences but contain no correct characters retrieved by baseline retrievers (e.g., ``\chinese{现汇交易具有高流动性}'' or ``\chinese{我国对外汇实行管制}''), and (2) randomly sampled correct sentences from the retrieval corpus. This approach enables \( M_{\text{Ret}} \) to learn error-tolerant retrieval patterns.

For a given correct-incorrect pair, assume there are $m$ correct sentences and $n$ negative samples. The fine-tuning loss is defined as:
$$
\mathcal{L} = -\frac{1}{m} \sum_{i=1}^m \log \frac{
  \exp\left( s(\mathbf{E}(x), \mathbf{E}(r_i^+)) / \tau \right)
}{
   \sum \exp\left( s(\mathbf{E}(x), \mathbf{E}(r_{i,j}^{+,-})) / \tau \right)
},
$$

where $\mathbf{E}(\cdot)$ denotes the embedding function of $M_{\text{Ret}}$, $s(\cdot, \cdot)$ is cosine similarity metric, $\{r_1^+, \dots, r_m^+\}$ are the positive samples retrieved from the target $y$, $\{r_1^-, \dots, r_n^-\}$ are the negative samples, and $\tau$ is a temperature parameter that controls the sharpness of the similarity distribution.

By fine-tuning $M_{\text{Ret}}$ in this manner, the model becomes more effective at handling noisy inputs, ensuring robust retrieval performance even in the presence of character-level errors.

\subsection{Multi-turn Length Reflection (MLR)}\label{sec:MLR}

To address the length constraints of the output, we introduce a mechanism that enables the model to "rethink" when the generated output length deviates from the expected target. The method for multi-turn reflection is illustrated in Algorithm \ref{alg:length_reflection}. Where $L(\cdot)$ denotes the operation to get the length of a sentence. The algorithm iteratively refines a corrected sentence using an LLM, ensuring its length matches the source sentence by repeatedly adjusting it through a template-based prompt until a predefined round limit is reached.

During each turn, we propose a length template as the length guidance for LLMs to analyze the output sentence. Notably, for different tasks, we collect lengths from the original input sentence and the output sentence to generate a length report which selects the proper length template from the task label, and then generates a sentence which compares the length of input $x$ and the output of $i_{th}$ round $y_i$, $T(x,y,task)$ denotes the length report of a sentence. The detailed length report could be found in Appendix \ref{apdx:length_template}.

This approach ensures that the model dynamically adjusts its responses when the lengths are unequal, enhancing both precision and adaptability.

\begin{algorithm}[!t]
\small
\caption{Multi-turn Length Reflection}
\label{alg:length_reflection}
\begin{algorithmic}[1]
\Require Source sentence $x$, first-round correction result $y_0$, round limit $m$, length template $T$, task label $task$
\Ensure Final corrected sentence $y$
\State $i \gets 0$
\While{$i < m$}
    \If{$T(x,y_i)$ satisfies length requirement}
        \State $y \gets y_i$
        \State \Return $y$
    \Else
        \State Compute $L(x)$ and $L(y_i)$
        \State Format $x$, $y_i$, $L(x)$ and $L(y_i)$ into prompt template $T$
        \State Generate length report as next-round prompt: $p \gets T(x, y_i, task)$
        \State Generate output using large language model: $y_{i+1} \gets \text{LLM}(p)$
    \EndIf
    \State $i \gets i + 1$
\EndWhile
\State $y \gets y_i$
\State \Return $y$
\end{algorithmic}
\end{algorithm}

\subsection{Adaptive Selection}

We propose an Adaptive Selection (AdSe) method to dynamically choose between retrieval-based and non-retrieval generation strategies based on task labels. If a training set is available, the retrieval-based method is prioritized; otherwise, the non-retrieval method is used by default. The system automatically switches to the alternative method if it fails to produce a valid sentence within $k$ iterations, ensuring robust correction. (See pseudo code in Appendix~\ref{apdx:MSC}).

\section{Experiments}
\begin{table}[!ht]
\centering
\small
\begin{tabular}{llrrr}
\toprule
Dataset & Domain & \#Sents & Len. & \# Errors \\
\midrule
\multirow{3}{*}{ECSpell Train} 
    & \textit{law} & 1,960 & 60.4 & 1,059 \\
    & \textit{med} & 3,000 & 99.5 & 1,473 \\
    & \textit{odw} & 1,728 & 81.7 & 1,000 \\
\cmidrule(lr){1-5}
\multirow{3}{*}{ECSpell Test}
    & \textit{law} & 500 & 58.5 & 255 \\
    & \textit{med} & 500 & 98.2 & 226 \\
    & \textit{odw} & 500 & 80.6 & 266 \\
\midrule
\multirow{7}{*}{LEMON}
    & \textit{car} & 3,245 & 85.9 & 1,577 \\
    & \textit{con} & 993 & 79.2 & 441 \\
    & \textit{enc} & 3,274 & 78.9 & 1,591 \\
    & \textit{gam} & 393 & 64.6 & 148 \\
    & \textit{mec} & 1,942 & 77.4 & 905 \\
    & \textit{new} & 5,887 & 49.3 & 2,941 \\
    & \textit{nov} & 5,982 & 71.5 & 3,006 \\
\midrule 

Aishell-1 Train & \textit{news} & 120,099 & 14.41 & 85,504 \\
\cmidrule(lr){1-5}
Aishell-1 Test & \textit{news} & 7,176 & 14.60 & 5,433 \\
\midrule
\multirow{2}{*}{CSEC Train}
    & \textit{news} & 7,869 & 30.7 & 8,344 \\
    & \textit{social} & 6,330 & 28.8 & 6,401 \\
\cmidrule(lr){1-5}
\multirow{2}{*}{CSEC Test}
    & \textit{news} & 1,174 & 30.9 & 1,268 \\
    & \textit{social} & 1,108 & 28.6 & 1,128 \\

\bottomrule
\end{tabular}
\caption{Detailed statistics of datasets used in the experiments.}
\label{tab:rag_data_stats}
\end{table}

\begin{table*}[!t]
    \centering
   \small
    \begin{tabular}{lcccccccccc}
        \toprule
        \multirow{2}{*}{Methods} & \multicolumn{3}{c}{ECSpell} & \multicolumn{7}{c}{LEMON} \\
        \cmidrule(lr){2-4} \cmidrule(lr){5-11}
         & \textit{law} & \textit{med} & \textit{odw} & \textit{car} & \textit{cot} & \textit{enc} & \textit{gam} & \textit{mec} & \textit{new} & \textit{nov} \\
        \toprule
        MacBERT & 38.6 & 28.7 & 37.7 & 31.1 & 44.2 & 28.8 & 12.9 & 40.7 & 29.7 & 12.6 \\
        BERT-MFT & 76.1 & 58.0 & 59.2 & 52.8 & 64.4 & 45.6 & 33.9 & 51.5 & 57.0 & 36.7 \\
        RELM & \textbf{91.2} & \textbf{82.4} & 83.6 & 53.6 & 67.7 & 47.7 & 34.6 & 53.9 & 58.8 & 38.0 \\
        \midrule
        GPT-3.5 & 38.0 & 26.0 & 47.4 & 21.4 & 30.7 & 29.6 & 11.8 & 29.8 & 22.8 & 15.2 \\
        GPT-3.5+\textbf{RAIR} & 45.4 & 31.9 & 54.0 & 27.0 & 40.9 & 33.8 & 14.2 & 32.0 & 25.0 & 16.7 \\
        Qwen2.5 & 46.9 & 23.5 & 54.4 & 26.6 & 33.5 & 34.7 & 20.8 & 24.3 & 34.6 & 24.2 \\
        Qwen2.5+\textbf{RAIR} & 68.0 & 40.5 & 73.0 & 38.8 & 50.1 & 49.4 & 27.1 & 47.9 & 45.2 & 34.1 \\
        DeepSeek-V3 & 82.7 & 67.6 & 90.2 & 56.2 & 65.6 & 60.6 & 40.9 & 68.7 & 68.8 & 57.7 \\
        DeepSeek-V3+\textbf{RAIR} & 87.4 & 76.6 & \textbf{92.0} & \textbf{60.4} & \textbf{69.1} & \textbf{61.6} & \textbf{43.7} & \textbf{70.6} & \textbf{70.5} & \textbf{59.4} \\
        \bottomrule
    \end{tabular}
     \caption{$F_1$-score performance comparison of different methods in ECSpell and LEMON datasets. The best results are marked in \textbf{bold}.}
     \label{tab:merged_f1_comparison_no_abla}
\end{table*}

\subsection{Datasets}

\textbf{Domain Spelling error} We select the following datasets, ECSpell \citep{lv_general_2023} and LEMON \citep{wu2023rethinking}, where ECSpell is a CSC benchmark dataset comprising three domains: legal (\textit{law}), medical (\textit{med}), and official document writing (\textit{odw}), with artificially constructed errors. LEMON, on the other hand, is a real-world Chinese spelling error dataset spanning seven domains: automotive (\textit{car}), contracts (\textit{cot}), encyclopedia (\textit{enc}), gaming (\textit{gam}), medicine (\textit{mec}), news (\textit{new}), and novels (\textit{nov}). Each dataset contains parallel sentence pairs where the source sentence contains errors and the target sentence provides the corrected version. While ECSpell contains a training dataset, LEMON does not include a training set, which serves to measure the generalizability of different CSC methods under a zero-shot setting. To fairly evaluate CSC's ability, we remove the sentence pair in the LEMON dataset whose source sentence length is unequal to the target sentence length. 

\noindent \textbf{ASR N-best Error} As for ASR N-best error correction, we select the ChineseHP/Aishell-1 dataset \citep{tang_pinyin_2024}, a specialized benchmark for news reading speech ASR correction. Each sentence contains the top-10 decoding result using the whisper model\footnote{\url{https://huggingface.co/BELLE-2/Belle-distilwhisper-large-v2-zh}}.

\noindent \textbf{Splitting Error} For splitting errors, we select the CSEC dataset proposed by \citep{liang-etal-2024-hybrid}, including machine-generated splitting characters based on a Chinese splitting dictionary on the social media and news domains to simulate the splitting online and during OCR. The detailed statistics of processed data are listed in Table \ref{tab:rag_data_stats}.

\subsection{Baselines}

\noindent \textbf{GPT-3.5}  is an enhanced version of GPT-3 developed by OpenAI, excelling in NLP tasks.

\noindent \textbf{Qwen2.5} \cite{qwen2025qwen25technicalreport} is a language model developed by Alibaba with strong language comprehension and multimodal capabilities.

\noindent \textbf{DeepSeek-V3} \cite{deepseekai2025deepseekv3technicalreport} is a foundation model from the DeepSeek team, designed for multitasking across coding, mathematics, reasoning, and multilingual processing. User inputs are routed to specialized experts within the model. For the LLM baselines, we use \textit{gpt-3.5-turbo}\footnote{\url{https://platform.openai.com/docs/models}}, \textit{qwen-plus}\footnote{\url{https://help.aliyun.com/zh/model-studio/developer-reference/what-is-qwen-llm}} and \textit{deepseek-chat}\footnote{\url{https://api-docs.deepseek.com/}} for the tasks. 

PLM baselines are selected for comparison.

\textbf{MacBERT} \cite{cui-etal-2020-revisiting} propose novel pretrain tasks for CSC,  adding mask tokens and using a transformer structure to predict the mask.

\textbf{T5} \cite{10.5555/3455716.3455856} proposes a unified framework, combining all text-based language problems into a text-to-text format.

\textbf{ReLM} \cite{wu2023rethinking} proposes rephrasing the language model and regards spelling correction as a rephrasing task. 

\subsection{Evaluation Metrics}
For spelling and splitting correction, we adopt the sentence-level criterion from \cite{wu2023rethinking}, where a correction $\hat{y}$ is considered correct if and only if it matches the reference $y$, reporting precision ($P$), recall ($R$), and $F_1 = 2*P*R/(P+R)$.

For ASR $N$-best correction, we implement the evaluation framework of \cite{tang_pinyin_2024}, calculating Character Error Rate (CER), and $\text{CER}=D(y,\hat{y})/L(y)$,
where $D(\cdot,\cdot)$ denotes the edit distance between hypothesis $\hat{y}$ and reference $y$, and $L(y)$ is the reference length. We compute Character Error Rate Reduction (CERR) for each model defined as follows:

$$
\text{CERR} = \frac{\text{CER}_{\text{baseline}} - \text{CER}_{\text{improved}}}{\text{CER}_{\text{baseline}}}
$$
\subsection{Detail Settings}
To improve the retriever's robustness to error-prone inputs, we fine-tune \textit{bge-m3} using the FlagEmbedding framework\footnote{\url{https://huggingface.co/BAAI/bge-m3}}, with a negative sampling size $k = 5$, learning rate $lr = 10^{-5}$, warm-up ratio of 0.1, contrastive temperature $\tau = 0.2$, and two training epochs. For ASR N-best error correction, the top 5 predictions are selected. We use the Pinecone vector database\footnote{\url{https://www.pinecone.io/}} to construct retrieval services. All experiments are conducted on two NVIDIA 3090 GPUs.

\subsection{Main Results}

\textbf{Domain Spelling Error}
Table~\ref{tab:merged_f1_comparison_no_abla} summarizes the $F_1$ scores across the ECSpell and LEMON datasets. On ECSpell, ReLM achieves the highest performance in the \textit{law} and \textit{med} domains, likely due to the similarity between its training data and the test distribution. However, DeepSeek-V3+RAIR outperforms all other methods in the \textit{odw} domain of ECSpell and across all domains of LEMON. The discrepancy in ReLM's performance—strong on ECSpell but weaker on LEMON—suggests that its reliance on training data limits generalization to real-world errors. Unlike ECSpell, where errors are artificially constructed, LEMON contains naturally occurring mistakes, highlighting the challenge of adapting to unseen error patterns.

Notably, all evaluated LLMs operate in a zero-shot setting, demonstrating their ability to correct spelling errors without task-specific fine-tuning. Among them, DeepSeek-V3 consistently surpasses GPT-3.5 and Qwen2.5 on both datasets. Furthermore, integrating RAIR improves performance across all LLM baselines, with the most significant gains observed for Qwen2.5. We hypothesize that Qwen's initial unequal length prediction provides a greater opportunity for RAIR's MLR module to refine. These results underscore the robustness of our framework in real-world scenarios, particularly in zero-shot settings where labelled training data is scarce or domain-mismatched.

\textbf{ASR N-best Error}  The experimental results are presented in Table \ref{tab:nbest_baseline}. The baseline performance using PLMs was originally reported by \cite{liang2024perlpinyinenhancedrephrasing}. Our experimental results demonstrate that modern LLMs, particularly Qwen2.5 and DeepSeek-V3, achieve significantly lower CERs compared to traditional PLM approaches. While GPT-3.5 initially exhibits higher CER values due to its weaker instruction-following capabilities in the zero-shot setting, our proposed RAIR framework effectively mitigates this limitation, reducing the CER by 35.9\%. Notably, the combination of capable instruction-following LLMs with RAIR yields the best performance, with DeepSeek-V3+RAIR achieving a remarkable 28.9\% relative improvement over the initial ASR decoded results.

\begin{table}[!h]
\centering

\small
\begin{tabular}{@{}llcc@{}}
\toprule
\textbf{Method} & \textbf{Type} & \textbf{CER\%$\downarrow$} & -CERR\%$\downarrow$ \\
\midrule
Top@1 & PLM & 5.84 & -- \\
MacBERT & PLM & 5.06 & \textcolor{teal}{-13.4} \\
T5 & PLM & 5.49 & \textcolor{teal}{-6.0} \\
ReLM & PLM & 5.20 & \textcolor{teal}{-11.0} \\
\midrule
GPT-3.5 & LLM & 9.84 & \textcolor{gray}{+68.5} \\
+\textbf{RAIR} & LLM & 6.31 & \textcolor{gray}{+8.1} \\
Qwen2.5 & LLM & 5.17 & \textcolor{teal}{-11.5} \\
+\textbf{RAIR} & LLM & 4.37 & \textcolor{teal}{-25.2} \\
DeepSeek-V3 & LLM & 4.94 & \textcolor{teal}{-15.4} \\
+\textbf{RAIR} & LLM & \textbf{4.15} & \textbf{\textcolor{teal}{-28.9}} \\
\bottomrule
\end{tabular}
\caption{Performance comparison on Aishell-1 dataset, where Top@1 indicates the best output decoded by the ASR model. The best result is marked in \textbf{bold}.}
\label{tab:nbest_baseline}
\end{table}

\textbf{Splitting Error}
\begin{table}[!tbp]
    \centering
    \small
    \begin{tabular}{lp{0.55cm}p{0.55cm}p{0.55cm}p{0.55cm}p{0.55cm}p{0.55cm}}
        \toprule
        \multirow{2}{*}{Methods} & \multicolumn{3}{c}{\textit{News}} & \multicolumn{3}{c}{\textit{Social}} \\
        \cmidrule(lr){2-4} \cmidrule(lr){5-7}
         & $P$ &$R$ & $F_1$ & $P$ & $R$ & $F_1$ \\

        \toprule

        Seq2Seq & 27.1 & 30.5 & 27.7 & 15.9 & 22.8 & 16.9 \\
        BERT & 70.5 & \textbf{75.9} & \textbf{73.1} & 44.1 & \textbf{46.6} & 45.3  \\
        \midrule
        GPT-3.5 & 13.6 & 13.6 & 13.6 & 5.6 & 6.1 & 5.8 \\
        +\textbf{RAIR} & 41.7 & 19.1 & 26.2 & 37.5 & 9.2 & 14.8 \\
        Qwen2.5 & 33.7 & 35.0 & 34.3 & 12.6 & 14.8 & 13.6 \\
        +\textbf{RAIR} & 56.4 & 40.5 & 47.2 & 38.7 & 20.6 & 26.9 \\
        DeepSeek-V3 & 54.4 & 53.9 & 54.2 & 57.7 & 39.6 & 47.0 \\
       +\textbf{RAIR} & \textbf{73.1} & 60.4 & 66.1 & \textbf{61.8} & 44.9 & \textbf{52.0} \\
        \bottomrule
    \end{tabular}
    \caption{Performance comparison of LLMs on CSEC dataset, where the best results are marked in \textbf{bold}. }
    \label{tab:performance_comparison_csec}
\end{table}
The performance of the CSEC split error correction dataset is shown in Table \ref{tab:performance_comparison_csec}. BERT and DeepSeek-V3+RAIR achieve the best results for CSEC-News and CSEC-Social, respectively, obtaining 73.1 and 52.0 in $F_1$ score, and DeepSeek-V3 greatly improves the precision of PLMs. We also observe a performance gap between the social and news datasets. This is because CSEC-News simulates the split of tokens in regular text, while CSEC-Social contains many informal social media expressions.  This disparity highlights the inherent challenge for LLMs in correcting non-standard linguistic constructions.


\begin{table}[!tbp]
\centering
\small

\begin{tabular}{llll}
\toprule
\multirow{2}{*}{Method} & \multicolumn{3}{c}{\textbf{\footnotesize CER\%}$\downarrow$ \footnotesize $_{-\text{CERR}\%\downarrow}$} \\ \cline{2-4} 
                & GPT-3.5 & Qwen2.5 & Deepseek-V3 \\
\midrule

Direct Correct   & 9.84    & 5.17 &   4.94   \\
\midrule
+RAIR         & 6.31 {\color{teal} $_{-35.9}$}   & 4.37{\color{teal} $_{-15.5}$} & 4.15{\color{teal} $_{-16.0}$}         \\
-\textit{w/o} Retrieval             & 9.36 {\color{teal} $_{-4.9}$}    & 4.40 {\color{teal} $_{-14.9}$}    &  4.20{\color{teal} $_{-14.9}$}          \\
-\textit{w/o} MLR        & 7.68{\color{teal} $_{-22.0}$}    & 4.81{\color{teal} $_{-7.0}$} &   4.67{\color{teal} $_{-5.5}$}       \\
-\textit{w/o} AdSe & 6.64{\color{teal} $_{-32.5}$}    & 4.55{\color{teal} $_{-12.0}$} &   4.45 {\color{teal} $_{-9.9}$}       \\

\bottomrule
\end{tabular}
\caption{Ablation study of different methods using LLMs on ChineseHP/Aishell-1 dataset, where the CER and CERRs are listed. Where "Direct Correct" denotes the prediction by directly using LLM. }
\label{tab:rag_aishell1}
\end{table}
\begin{table*}[!ht]
    \centering
    \small
    \begin{tabular}{llccccccccccc}
        \toprule
        \multirow{2}{*}{Base Model} & \multirow{2}{*}{strategy} & \multicolumn{3}{c}{ECSpell} & \multicolumn{7}{c}{LEMON} \\
        \cmidrule(lr){3-5} \cmidrule(lr){6-12}
        & & \textit{law} & \textit{med} & \textit{odw} & \textit{car} & \textit{cot} & \textit{enc} & \textit{gam} & \textit{mec} & \textit{new} & \textit{nov} \\
        \midrule
        \multicolumn{1}{c}{\multirow{4}{*}{DeepSeek-V3}} 
            & RAIR & 87.4 & 76.6 & 92.0 & 60.4 & 69.1 & 61.6 & 43.7 & 70.6 & 70.5 & 59.4 \\

            & \textit{w/o} Retrieval & 84.0 & 71.1 & 90.7 & 59.7 & 67.6 & 61.0 & 42.4 & 70.0 & 70.2 & 59.2 \\

            & \textit{w/o} MLR & 85.5 & 73.0 & 88.4 & 49.9 & 56.6 & 54.3 & 42.1 & 56.1 & 55.7 & 50.0 \\ 
            & \textit{w/o} AdSe & 84.5 & 71.7 & 91.9 & 53.0 & 60.4 & 58.5 & 41.2 & 59.4 & 56.9 & 51.9 \\


        \bottomrule
    \end{tabular}
    \caption{Ablation study of Deepseek-V3 on ECSpell and LEMON datasets}
    \label{tab:abla_csc}
\end{table*}
\subsection{Ablation Study}

The ablation experiment, which investigates the contributions of different submodules of the RAIR framework, is organised as follows: 1) Removing retrieval, 2) Removing MLR, 3) Removing adaptive selection

\textbf{ASR N-best Error} The ablation study for ChineseHP/Aishell-1 dataset is presented in Table \ref{tab:rag_aishell1}. Removing each submodule would result in a CER increase compared to RAIR.

\textbf{Domain Spelling Error} The ablation study for domain spelling error datasets ECSpell and LEMON is shown in Table \ref{tab:abla_csc}. The table demonstrates that the removal of the retrieval module and adaptive selection module would degrade the performance of LLMs. While removing MLR leads to a significant drop, we believe that because the evaluation criteria are rigid, only exact matches are regarded as correct.

\textbf{Splitting Error} The ablation study results on the CSEC splitting error dataset (Table \ref{tab:csec_abla}) reveal several key findings. Most notably, the removal of MLR causes the most substantial performance degradation in CSEC-News, primarily due to its critical role in maintaining output length constraints.

\begin{table}[!tbp]
    \centering
    \small
    \begin{tabular}{lp{0.55cm}p{0.55cm}p{0.55cm}p{0.55cm}p{0.55cm}p{0.55cm}}
        \toprule
        \multirow{2}{*}{Methods} & \multicolumn{3}{c}{\textit{News}} & \multicolumn{3}{c}{\textit{Social}} \\
        \cmidrule(lr){2-4} \cmidrule(lr){5-7}
         & $P$ &$R$ & $F_1$ & $P$ & $R$ & $F_1$ \\

        \toprule


        DeepSeek-V3 & 54.4 & 53.9 & 54.2 & 57.7 & 39.6 & 47.0 \\
        \midrule
                +RAIR & 73.1 & 60.4 & 66.1 & 61.8 & 44.9 & 52.0 \\
        
        -\textit{w/o} Retrieval & 55.8 & 54.2 & 55.0 & 58.3 & 39.6 & 47.2 \\
        -\textit{w/o} MLR & 53.1  & 48.5 & 50.7 & 56.1 & 40.5 & 47.0 \\
        -\textit{w/o} AdSe & 71.3 & 57.7 & 63.8 & 60.4 & 42.6 & 49.9 \\

        \bottomrule
    \end{tabular}
    \caption{Ablation Study of DeepSeek on CSEC dataset}
    \label{tab:csec_abla}
\end{table}

\subsection{Analysis}

\textbf{Multi-turn Length Reflection} DeepSeek's superior length correction capability is shown in Table~\ref{tab:results_sentence_correction}. The MLR module improves DeepSeek's correct length prediction samples from (494,467,492) to (500,489,495), while Qwen shows significant MLR-driven improvements (+141,+146,+98), consistent with its documented instruction-following strengths \citep{dong2024self}. This demonstrates that the MLR module could generate better results, subject to length requirements, improving correction performance.
\begin{table}[htbp]
\small
\centering

\begin{tabular}{lrrrr}
\toprule
\multirow{2}{*}{Model} & \multicolumn{3}{c}{Domain} & \multirow{2}{*}{Total} \\
\cmidrule(lr){2-4}
 & Law & Med & Odw & \\
\midrule
GPT3.5 & 438 & 362 & 447 & 1247 \\
GPT3.5+MLR & 464 & 399 & 464 & 1327 \\
Qwen2.5 & 337 & 231 & 322 & 890 \\
Qwen2.5+MLR & 478 & 377 & 420 & 1275 \\
DeepSeek-V3 & 494 & 467 & 492 & 1453 \\
DeepSeek-V3+MLR & \textbf{500} & \textbf{489} & \textbf{495} & \textbf{1484} \\
\midrule
\#Sents & 500 & 500 & 500 & 1500 \\
\bottomrule
\end{tabular}

\caption{Correct length statistics using different models in ECSpell dataset, where the number indicates the number of predictions which have equal length with the source sentence, best results are marked in \textbf{bold}}
\label{tab:results_sentence_correction}
\end{table}



\paragraph{Retrieval Module}
To demonstrate the effectiveness of our proposed fine-tuned retriever method, we evaluate its performance on the ECSpell dataset using Hit@5 and Mean Reciprocal Rank (MRR), as shown in Table \ref{tab:hit@5_performance} and Table \ref{tab:mrr_performance}. A hit is recorded if the correct Chinese character appears in the retrieved candidates, and MRR measures how the retriever ranks the first relevant sentence for a set of queries. Our fine-tuned \textit{bge-m3} retriever achieves higher relevance in candidate generation, providing more accurate contextual information for LLMs in subsequent error correction steps.

\begin{table}[htbp]
\centering

\small
\begin{tabular}{lccc}
\toprule

\multirow{2}{*}{Method} & \multicolumn{3}{c}{Hit@5$\uparrow$} \\ \cline{2-4}
 & \textit{law} & \textit{med} & \textit{odw} \\
\midrule
\#Errors & 390 & 356 & 407 \\
bge-m3 & 315 & 185 & 328 \\
bge-m3-ft & \textbf{369} & \textbf{248} & \textbf{352} \\

\bottomrule
\end{tabular}
\caption{Hit@5 comparison of retrieval methods in ECSpell test data, where \#Errors indicates the total number of error characters.}
\label{tab:hit@5_performance}
\end{table}
\begin{table}[htbp]
\centering

\small
\begin{tabular}{lccc}
\toprule
\multirow{2}{*}{Method} & \multicolumn{3}{c}{\footnotesize MRR$\uparrow$} \\ \cline{2-4}
 & \textit{law} & \textit{med} & \textit{odw} \\
\midrule
bge-m3 & 0.87 & 0.76 & 0.85 \\
bge-m3-ft & 0.88 & 0.83 & 0.90 \\
\bottomrule
\end{tabular}
\caption{MRR comparison of retrieval methods in ECSpell test data.}
\label{tab:mrr_performance}
\end{table}


\section{Conclusion}

This work proposes \textbf{RAIR}, a plug-and-play Retrieval-Augmented Generation (RAG) framework for error correction featuring multi-turn length reflection. The framework dynamically constructs a retrieval corpus by combining training data with LLM-generated synthetic samples, then fine-tunes the semantic search retriever to catch the error correction pattern. Through iterative multi-turn length reflection and adaptive selection, RAIR can handle fixed and variable-length correction scenarios while enforcing output length constraints. Designed for model-agnostic deployment, the framework demonstrates consistent performance improvements across diverse LLM backbones, achieving superior correction accuracy compared to baseline models.

\section*{Limitations}

Domain Chinese Spelling Correction requires specialized domain knowledge integration. Our framework successfully incorporates this knowledge through an RAG approach. Building on this foundation, future research directions include enhancing integration of phonetic and glyph features within the RAG architecture and systematic investigation of prompt engineering strategies through iterative optimization to identify optimal prompting configurations.

\bibliography{anthology,custom}
\bibliographystyle{acl_natbib}

\clearpage
\appendix

\section{LLM prompts for error correction}
\label{sec:appendix}
In this section, we provide specific prompts for spelling, splitting and ASR N-best error correction. We include the original Chinese prompts and the corresponding English translation. Note that in the experiment we only use Chinese prompt. 

\begin{tcolorbox}[
    breakable,
    colback=blue!8,
    title=Spelling Errors,
]
\small

\chinese{你是一位著名的语言学家，请你纠正一下句子中汉字拼写的错误，你需要识别并纠正用户输的句中可能的错别字并输出正确的句子，纠正时必须保证改动前后句必须等长，在纠正错别字的同时尽可能减少对原句的改动（不添加额外标点符号，不添加额外的字，不删除多余的字）。只输出没有错别字的句子，不要添加任何其他解释或说明。如果句子没有错别字，就直接输出和输入相同的句子。 现在请纠正下列句子：<\text{input\_sentence}>}

(You are a renowned linguist. Please correct any Chinese character spelling errors in the sentence. You need to identify and correct possible wrong/misused characters in the user's input while ensuring the corrected sentence maintains the same length as the original. Minimize changes to the original sentence (no extra punctuation, no additional characters, no deletion of excess characters). Output only the corrected sentence without any additional explanation. If there are no errors, output the same sentence as input. Now please correct the following sentence: <\text{input\_sentence}>)

\end{tcolorbox}

\begin{tcolorbox}[
    breakable,
    colback=blue!8,
    title=Splitting Errors,
]
\small

\chinese{你是一位著名的语言学家，请你纠正一下句子中汉字拆分的错误，你需要识别并纠正用户输的句中可能的拆分字并输出正确的句子，纠正时必须保证被拆开的汉字得到还原，在纠正拆分字的同时尽可能减少对原句的改动。因此你的纠正中，输出句子的长度小于等于输入句子的长度。只输出没有错别字的句子，不要添加任何其他解释或说明。如果句子没有拆分字，就直接输出和输入相同的句子。 现在请纠正下列句子：<\text{input\_sentence}>}

(You are a renowned linguist. Please correct any Chinese character splitting errors in the sentence. You need to identify and restore any incorrectly split characters while minimizing changes to the original sentence. The output sentence length should be less than or equal to the input length. Output only the corrected sentence without explanation. If no splitting errors exist, output the original sentence. Now please correct the following sentence: <\text{input\_sentence}> )

\end{tcolorbox}

\begin{tcolorbox}[
    breakable,
    colback=blue!8,
    title=N-best Errors,
]
\small

\chinese{你是一位著名的语言学家，请你根据以下语音识别中的多候选句子进行纠正，你需要识别并纠正多候选的句子中可能的错别字并输出正确的句子，多候选的每个句子用｜分割，为了完成这一目标，你需要比较多个候选，确定正确句子的长度，并对错别字进行纠正，因此你的纠正中，输出句子的长度可能大于某个输入句子的长度，也可能小于某个输入句子的长度，也可能等于某个输入句子的长度。只输出最后识别的句子，不要添加任何其他解释或说明。 现在请纠正下列句子：<\text{input\_sentence}> }

(You are a renowned linguist. Please correct errors in these speech recognition N-best candidates (separated by |). Compare the candidates to determine the correct sentence length and correct any wrong characters. The output length may be greater than, less than, or equal to any input candidate's length. Output only the final identified sentence without any additional explanation. Now please correct the following sentence: <\text{input\_sentence}>)

\end{tcolorbox}

\section{Length Templates}\label{apdx:length_template}

In our experimental framework, we design length templates to analyze three distinct error types: spelling errors, splitting errors, and N-best errors. For spelling and splitting errors, the correct sentence length can be precisely determined from the reference text. However, for N-best errors where the exact correct length is inherently ambiguous, we adopt a conservative estimation approach by bounding the potential correct length between the longest and shortest decoding results in the N-best list. This methodology provides a principled way to handle length variation in error analysis while accounting for the uncertainty in N-best hypotheses.

\begin{tcolorbox}[breakable, colback=yellow!8, title=Length Requirements Satisfied]
\small

\chinese{源句 $x$ 和输出句 $y$ 的长度符合任务要求！}

(The lengths of the source sentence $x$ and the output sentence $y$ satisfy the requirements of the task)

\end{tcolorbox}

\begin{tcolorbox}[breakable, colback=yellow!8, title=Spelling Errors]
\small

\chinese{源句 $x$ 和输出句 $y$ 长度不相等。源句有 $L(x)$ 个字符，输出句有 $L(y)$ 个字符，请重新纠正！}

(The lengths of the source sentence $x$ and the output sentence $y$ are not equal. The source sentence has $L(x)$ characters, and the output sentence has $L(y)$ characters. Please correct it again!)

\end{tcolorbox}

\begin{tcolorbox}[breakable, colback=yellow!8, title=Splitting Errors]
\small

\chinese{输出句 $y$ 的长度超过源句 $x$ 的长度。源句有 $L(x)$ 个字符，输出句有 $L(y)$ 个字符，请确保输出句长度不超过源句长度！}

(The length of the output sentence $y$ exceeds the length of the source sentence $x$. The source sentence has $L(x)$ characters, and the output sentence has $L(y)$ characters. Please ensure the output sentence length does not exceed the source sentence length!)

\end{tcolorbox}

\begin{tcolorbox}[breakable, colback=yellow!8, title=N-best Errors]
\small

\chinese{输出句 $y$ 的长度不在候选列表 $X$ 的最小长度和最大长度之间。候选列表 $X$ 的最小长度为 $\min(L(X))$，最大长度为 $\max(L(X))$，输出句有 $L(y)$ 个字符，请确保输出句长度在候选列表的长度范围内！}

(The length of the output sentence $y$ is not within the range of the minimum and maximum lengths of the candidate list $X$. The minimum length of the candidate list $X$ is $\min(L(X))$, the maximum length is $\max(L(X))$, and the output sentence has $L(y)$ characters. Please ensure the output sentence length falls within the length range of the candidate list!)

\end{tcolorbox}

\section{Multi-source Correction Algorithm}\label{apdx:MSC}
The pseudo code of Multi-source Correction is shown in algorithm \ref{alg:multi_source_correction}.
\begin{algorithm}
\caption{Multi-Source Correction}
\label{alg:multi_source_correction}
\begin{algorithmic}[1]
\Require Sentence with typos $x$, task label $\text{TaskType}$, maximum iterations $k$
\Ensure Corrected sentence $y$

\State Initialize $\text{counter} \gets 0$
\State $\text{Method}_1 \gets \text{Retrieval-based Correction}$, $\text{Method}_2 \gets \text{Non-retrieval Correction}$
\If{$\text{TaskType}$ does not contain training set}
    \State Swap $\text{Method}_1$ and $\text{Method}_2$
\EndIf

\State $y \gets \text{Method}_1(x)$ \Comment{First apply $\text{Method}_1$ to generate initial result}
\If{$L(x) \neq L(y)$}
    \State $\text{counter} \gets \text{counter} + 1$
    \If{$\text{counter} \geq k$}
        \State $y \gets \text{Method}_2(x)$ \Comment{Switch to $\text{Method}_2$ for correction}
        \State $\text{counter} \gets 0$
    \EndIf
\EndIf

\State \Return $y$
\end{algorithmic}
\end{algorithm}

\begin{table}[htbp]
\footnotesize
\centering

\begin{tabular}{lp{5cm}}
\toprule

\textbf{Example 1} & \chinese{建设用地使用券自登记机构登记时设立,不经登记不生}\textcolor{red}{\chinese{肖}}\chinese{力} \\ 
\textbf{DeepSeek} & \chinese{建设用地使用权自登记机构登记时设立,不经登记不生}\textcolor{blue}{\chinese{效}}\textcolor{orange}{\#} \\
\textbf{DeepSeek+MLR} & \chinese{建设用地使用权自登记机构登记时设立,不经登记不生}\textcolor{blue}{\chinese{效}}\chinese{力} \\

\midrule

\textbf{Example 2} & \chinese{根据}\textcolor{red}{\chinese{起夜}}\chinese{破产法规定, }\textcolor{red}{\chinese{宅}}\chinese{权人会议成员中,下列人享有表决权的有} \\ 
\textbf{DeepSeek} & \chinese{根据}\textcolor{blue}{\chinese{企业}}\chinese{破产法规定, }\textcolor{blue}{\chinese{债}}\chinese{权人会议成员中,下列人}\textcolor{orange}{\chinese{员}}\chinese{享有表决权的有}\\
\textbf{DeepSeek+MLR} & \chinese{根据}\textcolor{blue}{\chinese{企业}}\chinese{破产法规定, }\textcolor{blue}{\chinese{债}}\chinese{权人会议成员中,下列人享有表决权的有} \\

\midrule

\textbf{Example 3} & \chinese{食品安全抽检覆盖}\textcolor{red}{\chinese{王}}\chinese{部食品类别、品种}  \\ 
\textbf{DeepSeek} & \chinese{食品安全抽检覆盖}\textcolor{orange}{\chinese{完整}}\chinese{部食品类别、品种}\\
\textbf{DeepSeek+Ret.} & \chinese{食品安全抽检覆盖}\textcolor{blue}{\chinese{全}}\chinese{部食品类别、品种} \\

\midrule

\textbf{Example 4} & \chinese{这里松柏长青,修竹滴翠,溪水流香,潭似青砚,洞深莫测,真是风景这边独好,娘娘顶上建}\textcolor{red}{\chinese{逢}}\chinese{莱。} \\ 
\textbf{DeepSeek} & \chinese{这里松柏}\chinese{\textcolor{orange}{常}}\chinese{青,修竹}\chinese{\textcolor{orange}{翠绿}}\chinese{,溪水}\chinese{\textcolor{orange}{清}}\chinese{香,潭}\chinese{\textcolor{orange}{如}}\chinese{青砚,洞深莫测,}\chinese{\textcolor{orange}{风景独特}}\chinese{,娘娘顶上建}\chinese{\textcolor{orange}{有}}\chinese{\textcolor{blue}{蓬}}\chinese{莱。}\\
\textbf{DeepSeek+Ret.} & \chinese{这里松柏长青,修竹滴翠,溪水流香,潭似青砚,洞深莫测,真是风景这边独好,娘娘顶上建}\textcolor{blue}{\chinese{蓬}}\chinese{莱。} \\

\bottomrule
\end{tabular}
\caption{Example of MLR and retrieval correction}
\label{tab:rag_case}
\end{table}

\section{Case Study}
Case studies are shown in Table \ref{tab:rag_case} where red characters indicate erroneous characters, orange characters indicate incorrect corrections, and blue characters indicate correct corrections. The symbol "\#" is added as a placeholder for demonstration purposes.

Examples 1 and 2 show that introducing the multi-round reflection mechanism can effectively correct length-related errors. For instance, "\chinese{生肖力}" ("zodiac force") should be corrected to "\chinese{生效力}" ("be effective"), but the model initially corrected it to "\chinese{生效}", failing to meet the length constraint. After multi-length reflection, "\chinese{生肖力}" was successfully corrected to "\chinese{生效力}". Similarly, in Example 2, the single-round correction mistakenly changed "\chinese{下列人}" ("the following persons") to "\chinese{下列人员}" ("the following personnel"), which was an unnecessary modification.

In Example 3, the sentence comes from the ECSpell dataset. The retrieval mechanism successfully corrected "\chinese{王部}" ("king's department") to "\chinese{全部}" ("all"). Example 4 features a sentence from the LEMON dataset's novel category containing elegant expressions like "\chinese{松柏长青}" ("pine trees stay evergreen") and "\chinese{修竹滴翠}" ("cultivated bamboo drips with emerald"), which the model correctly preserved without over-correction.

\section{Average correction rounds}
The multi-turn approach presents efficiency challenges due to API call costs. Our analysis in Table~\ref{tab:rounds_required} reveals the MLR module's effectiveness: over 80\% of Qwen's and 50\% of GPT-3.5's length errors are corrected in just one reflection round. This efficiency stems from our length reflection algorithm (Algorithm~\ref{alg:length_reflection}), which leverages both the initial erroneous prediction and target length to guide subsequent generations. Notably, we focus on cases solvable within four rounds, demonstrating the module's practical viability despite the inherent multi-turn overhead.

\begin{table}[h]
\small
\centering

\begin{tabular}{llcccc}
\toprule
Model & Domain & R1 & R2 & R3 & R4 \\
\midrule
\multirow{3}{*}{GPT3.5} 
    & \textit{law} & 13 & 8 & 4 & 1 \\
    & \textit{med} & 19 & 7 & 6 & 1 \\
    & \textit{odw} & 10 & 4 & 2 & 0 \\
\midrule
\multirow{3}{*}{Qwen2.5} 
    & \textit{law} & 112 & 19 & 7 & 3 \\
    & \textit{med} & 114 & 16 & 5 & 11 \\
    & \textit{odw} & 82 & 14 & 3 & 2 \\
\midrule
\multirow{3}{*}{Deepseek-V3} 
    & \textit{law} & 4 & 1 & 0 & 1 \\
    & \textit{med} & 18 & 4 & 2 & 1 \\
    & \textit{odw} & 2 & 0 & 1 & 0 \\
\bottomrule
\end{tabular}
\caption{Results across rounds for different models and domains}
\label{tab:rounds_required}
\end{table}

The observation on multi-turn introduces a challenge related to efficiency, as each API call incurs both time and financial costs. To further assess the effectiveness and efficiency of the MLR module, we present statistics on the desired length correction and the number of reflection rounds for each sentence pair, as shown in Table \ref{tab:rounds_required}. We focus on sentences where the length could be accurately corrected within four rounds using the MLR. For Qwen2.5, over 80\% of the incorrectly predicted sentence lengths could be corrected in just one round, while for GPT-3.5, more than 50\% are rectified within one round. This arises from integrating the erroneous initial length prediction and the target length derived from the source sentence in our length reflection module to guide the generation for the next round, as outlined in Algorithm \ref{alg:length_reflection}.


\end{document}